\documentclass{article}
\usepackage{spconf,graphicx,siunitx,microtype,amsfonts}
\usepackage{times}
\usepackage{soul}
\usepackage{url}
\usepackage[hidelinks]{hyperref}
\usepackage[utf8]{inputenc}
\usepackage[small]{caption}
\usepackage{amsmath}
\usepackage{amsthm}
\usepackage{booktabs}
\usepackage{algorithm}
\usepackage{algorithmic}
\usepackage{amsthm,amsmath,amssymb}
\usepackage{mathrsfs}
\usepackage{multirow}
\usepackage[switch]{lineno}
\usepackage{xcolor}
\usepackage{wrapfig}

\usepackage{subfigure}
\usepackage{array}
\usepackage{multirow}
\usepackage{booktabs}
\let\oldhat\hat

\renewcommand{\hat}[1]{\oldhat{\mathbf{#1}}}

\newcommand{\ie}{\emph{i.e., }}
\newcommand{\eg}{\emph{e.g., }}

\title{Joint Multi-Facts Reasoning Network for Complex Temporal Question Answering over Knowledge Graph}
\name{Rikui Huang$^{\spadesuit\clubsuit\blacklozenge}$, Wei Wei$^{\spadesuit\dagger}$\thanks{$^{\dagger}$ Corresponding author.} \thanks{This work was supported in part by the National Natural Science Foundation of China under Grant No. 62276110, and in part by the fund of Joint Laboratory of HUST and Pingan Property \& Casualty Research (HPL). }, Xiaoye Qu$^{\spadesuit}$
,Wenfeng Xie$^{\blacktriangle}$, Xianling Mao$^{\bigstar}$, Dangyang Chen$^{\blacktriangle}$
}
\address{
$^\spadesuit$ School of Computer Science \& Technology, Huazhong University of Science and Technology \\
$^\clubsuit$Institute of Artificial Intelligence, Huazhong University of Science and Technology \\
$^\blacklozenge$School of Artificial Intelligence \& Automation, Huazhong University of Science and Technology \\
$^\blacktriangle$Ping An Property \& Casualty Insurance Company of China, Ltd, Shenzhen, China\\
$^\bigstar$School of Computer Science \& Technology, Beijing Institute of Technology, Beijing, China
}

\begin{document}
%
\maketitle
\begin{abstract}
Temporal Knowledge Graph (TKG) is an extension of regular knowledge graph by attaching the time scope. Existing temporal knowledge graph question answering (TKGQA) models solely approach simple questions, owing to the prior assumption that each question only contains a single temporal fact with explicit/implicit temporal constraints. 
Hence, they perform poorly on questions which own multiple temporal facts. In this paper, we propose \textbf{\underline{J}}oint \textbf{\underline{M}}ulti \textbf{\underline{F}}acts \textbf{\underline{R}}easoning \textbf{\underline{N}}etwork (JMFRN), to jointly reasoning multiple temporal facts for accurately answering \emph{complex} temporal questions. Specifically, JMFRN first retrieves question-related temporal facts from TKG for each entity of the given complex question. For joint reasoning, we design two different attention (\ie entity-aware and time-aware) modules, which are suitable for universal settings, to aggregate  entities and timestamps information of retrieved facts. Moreover, to filter incorrect type answers, we introduce an additional answer type discrimination task. Extensive experiments demonstrate our proposed method significantly outperforms the state-of-art on the well-known complex temporal question benchmark TimeQuestions.
\end{abstract}
\section{Introduction}
\label{sec:intro}

Regular knowledge graph usually organizes facts in the form of triples like (\emph{subject}, \emph{relation}, \emph{object}) \cite{ji2021survey, li2023two,li2023performant}. However, knowledge is dynamically updated rather than stationary. For example, the fact \textit{(Donald Trump, hold position, President)} is valid only for a certain period of time, \eg [2016, 2020]. To portray temporal constraint on facts, temporal knowledge graph (TKGs) are proposed, where facts are associated with time, \ie (\emph{subject}, \emph{relation},\emph{object},[\emph{start time},\emph{end time}]). 

Different from conventional Knowledge Graph Question Answering (KGQA), Temporal Knowledge Graph Question Answering (TKGQA) aims to answering a natural question containing temporal constraints over TKGs. Recently, many efforts have been devoted to the TKGQA task and formulate it as a \textbf{temporal knowledge graph completion} problem, such as CRONKGQA~\cite{saxena2021question} and TempoQR~\cite{mavromatis2022tempoqr}. They project the contained entities, timestamps and relations into TKG embeddings and predict the missing part  (\eg entity \emph{or} timestamps) of the triple based on the existing facts over the given TKG. For example, a question  \textit{``Which movie won the Best Picture in 1973?"} to (\textit{Best Picture}, \textit{Won By}, ?, [1973, 1973]). Despite effective, these methods faces several problems: 1) Prior assumption that a question solely contains one entity; and 2) Prerequisite that additional entity-specific annotations (subject \emph{or} object) of the grounded entities are known in advance. These limitations constrain previous works in yielding correct answers towards \emph{complex} temporal questions.

\begin{figure*}[ht]
	\centering
	\includegraphics[width=1.5\columnwidth]{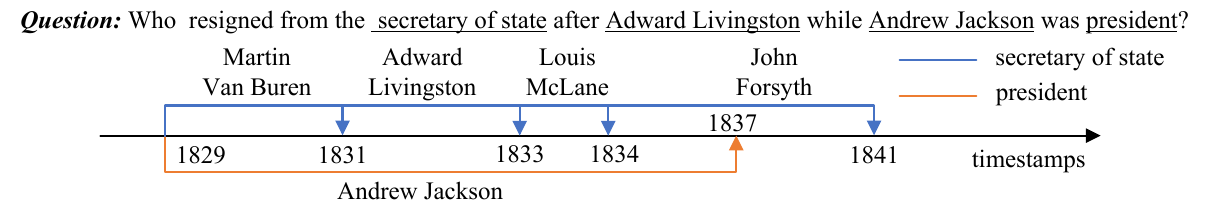}
	\caption{An example of a complex temporal question. The answer of the question is \textit{ Louis Mclane}.}
	\label{case}
\end{figure*}

Complex temporal questions are striking difference from simple ones, which usually contain multiple entities, and thus naturally increases the difficulty of the TKGQA task, as they commonly require jointly reasoning on multiple temporal facts derived from TKGs. Without loss of generality, we take the question in Figure~\ref{case} as an example for illustration, for answering the question \textit{``Who resigned from the secretary of state after Adward Livingston while Andrew Jackson was president?"} two temporal facts, \ie (\textit{Adward Livingston}, \textit{hold position}, \textit{secretary of state}, [1831, 1833]) and (\textit{Andrew Jackson}, \textit{hold position}, \textit{president}, [1829, 1837]), are required simultaneously, solely relying on either of which may lead to incorrect answer.

Reasoning over multi-hop on relation paths the KG is a strategy for joint multi-fact reasoning for traditional KGQA. However, it is still non-trivial for TKGQA since most temporal facts are temporally related instead of relational connected. To address the problem, in this paper we propose a Joint Multi-Facts Reasoning Network (JMFRN) to handle complex temporal questions by considering multiple facts simultaneously without additional entity-specific annotations. To combine information from multiple facts, we design two different attention modules to aggregate entity information and temporal information in retrieved facts respectively. Finally, the aggregated representations are applied for joint reasoning to obtain more confident answers. In addition, the answer to temporal questions generally can be categorized into two types, entity or timestamp. To correspond the answer type to the question, an answer type discrimination assistance task is designed to joint train with JMFRN.

To sum up, our contributions are summarized as follows. First, we make the effort to target the \emph{complex} question answering over TKGs in practice, and propose a simple but effective TKGQA model (JMFRN) to jointly reasoning with multiple temporal facts. The two well-designed attentions (\ie entity-aware and time-aware) modules for multiple temporal facts information aggregation are both proposed for improving the prediction accuracy. Second, to avoid irrelevant answers, we design an auxiliary answer type classification task for jointly training. In practice, we find that the auxiliary task not only improves accuracy but also accelerates and stabilises model training. Finally, we achieve the state-of-the-art performance on temporal complex questions dataset TimeQuestions with 0.628 Hits@1. In particular, for complex questions with multiple entities, our method presents convincing results.


\section{Related work}
\label{sec:related}

Recently, question answering (QA) \cite{wang2023dynamic, song2023answering, zhu2023mirror, BIBM} has attracted the attention of many researchers. Knowledge graph question answering (KGQA) is one of question answering tasks. Some simple questions can be answered accurately just by retrieving the relevant facts accurately. Furthermore, aimed at improving complex questions answering which requires multiple facts by expanding one-hop Q\&A \cite{sun2018open} to multi-hop Q\&A \cite{sun2019pullnet, huang2023mvp}. However, these methods neglect the fact that there are many questions with temporal constraints. 

Temporal knowledge graph question answering (TKGQA) focuses on reasoning in the time dimension \cite{chen2022temporal}. To model temporal information, CRONKGQA \cite{saxena2021question} and TempoQR \cite{mavromatis2022tempoqr} modeled entities and timestamps through temporal knowledge graph embedding (TKGE) \cite{lacroix2019tensor, gu2021read, gu2022delving, qu2023distantly}. TSQA \cite{shang2022improving} designed a timestamp estimation module to enhance the time sensitivity of the model. TMA \cite{TMA} improves Q\&A performance through enhanced fact retrieval and adaptive fusion network. Despite successful, they ignore the fact that complex temporal questions need to be answered jointly with multiple facts. On the other hand, EXAQT \cite{jia2021complex} encodes temporal facts via RNN and incorporates temporal information, which is not always available in practical scenarios, into traditional KGQA models. Our proposed JMFRN aims to perform complex temporal question answering by joint multiple facts without any priori entity position and temporal information.

\section{Preliminaries}
\label{sec:preliminaries}
\textbf{TKGQA.} Given a natural language question $q$ and a TKG $\mathcal{G}:=(\mathcal{E},\mathcal{R},\mathcal{T},\mathcal{F})$, TKGQA aims to find a suitable entity $e \in \mathcal{E}$ or timestamp $\tau \in \mathcal{T}$. Each fact $(s,r,o, \tau ) \in \mathcal{F}$ is a quadruple, where $s,o \in \mathcal{E}$ denote the subject and object entities, respectively, $r \in \mathcal{R}$ is the relation between the two entities, and $\tau \in \mathcal{T}$ is a timestamp.\\
\textbf{TKGE.} Given a TKG $\mathcal{G}=(\mathcal{E},\mathcal{R},\mathcal{T},\mathcal{F})$, TKGE learn $D$-dimensional vectors ${\bf e}_\varepsilon, {\bf v}_r, {\bf t}_\tau \in \mathbb{R}^D$ for $\varepsilon \in \mathcal{E}$, $r \in \mathcal{R}$ and $\tau \in \mathcal{T}$. Later, TKGE designs a score function $\varphi(\cdot)$ such that $\varphi({\bf e}_s, {\bf v}_r, {\bf e}_o, {\bf t}_\tau)>\varphi({\bf e}_{s'}, {\bf v}_{r'}, {\bf e}_{o'}, {\bf t}_{\tau'})$, where $(s,r,o,\tau) \in \mathcal{F}$ is a valid fact and $(s',r',o',\tau') \notin \mathcal{F}$ is an invalid fact. In this paper, we adopt TComplEx as the TKGE model. TComplEx \cite{lacroix2019tensor} is an extension of the ComplEx \cite{trouillon2016complex} designed for TKGs. TComplEx embeds elements such as entities into the complex space $\mathbb{C}^{D/2}$ and it designs the corresponding score function as follows $\varphi({\bf e}_s, {\bf v}_r, {\bf e}_o, {\bf t}_\tau)={\bf Re}(\left \langle{\bf e}_s, {\bf v}_r \odot {\bf t}_\tau, \overline{{\bf e}}_o\right \rangle)$, where ${\bf Re}(\cdot)$ denotes the real part, $\overline{{\bf e}}_o$ is the complex conjugate of ${\bf e}_o$ and $\odot$ is the element-wise product.

\section{Method} \label{section: Method}
\label{sec:method}

\begin{figure}[t]
	\centering
	\includegraphics[width=0.8\columnwidth]{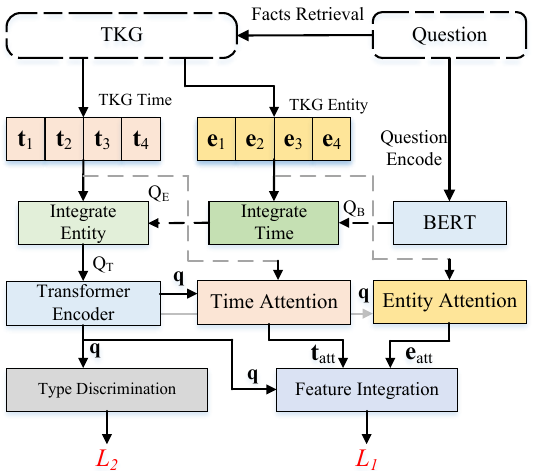}
	\caption{The overview of of our proposed JMFRN.
 }
	\label{framework}
\end{figure}


\subsection{Facts Retrieval}
Given a complex question, there are multiple entities $e_1,e_2,...,e_k$ in the question. To answer this question, we can retrieve facts based on these entities in a TKG to enhance the question representation. All question-related facts will provide individual time information for each entity, i.e. start time $t^s_1,t^s_2,...,t^s_k$ and end time $t^e_1,t^e_2,...,t^e_k$. By integrating the entities and the temporal information into the question representation, the model can make a better understanding of the question. Formally, for a temporal question text $\bf q$, we first obtain a representation of each token in BERT \cite{kenton2019bert}:
\begin{equation}
    {\bf Q}_B={\bf W}_B {\bf BERT}({\bf q}_{in})
\end{equation}
where ${\bf Q}_B := \left[{\bf q}_{B_{CLS}},{\bf q}_{B_1},...,{\bf q}_{B_N} \right]$ is a $D \times N$ embedding matrix, ${\bf W}_B$ is a $D \times D_B$ learnable projection matrix. Then, we project the entities and timestamps in questions separately:
\begin{equation}
{\bf Q}_{E_i}=\left\{
\begin{array}{lcl}
{\bf W}_E {\bf e}_{\varepsilon}       &      & \text{if token \textit{i} is an entity $\varepsilon$}\\
{\bf W}_T {\bf e}_{\tau}     &      & \text{if token \textit{i} is a timestamp $\tau$}\\
{\bf Q}_{B_i}     &      & \text{otherwise.}\\
\end{array} \right. 
\end{equation}
where ${\bf W}_E$ is a $D \times D$ learnable projection. To further integrate more temporal information into the question representation, we fuse entity information with corresponding time information.
\begin{equation}
{\bf Q}_{T_i}=\left\{
\begin{array}{lcl}
{\bf Q}_{E_i}+{\bf W}_{st}{\bf t}^s_i+{\bf W}_{et}{\bf t}^e_i     &    & \text{if token \textit{i} is an entity}\\
{\bf Q}_{E_i}       &      & \text{otherwise}\\
\end{array} \right.
\end{equation}
where ${\bf W}_{st}$ and ${\bf W}_{et}$ are two projection matrices to distinguish start and end timestamps. After $l$-layer transformers encoding, we take [CLS] as the representation of the question.

\subsection{Entity-aware/Time-aware Attention}
A complex temporal question may contain multiple entities without annotations. Selecting a correct entity as a query entity will help the model to get the correct answer. To jointly reasoning with multiple facts, we design an Entity-aware Attention Module (EAM) and a Time-aware Attention Module (TAM) to aggregate information from all retrieved facts.
\begin{equation}
{\bf e}_{att}= \sum\nolimits^k_{i=1} \alpha_{i}{\bf e}_i , \quad \quad  {\bf t}_{att}= \sum\nolimits^k_{i=1} \beta_{i}{\bf t}_i
\end{equation}
\begin{equation}
\alpha_{i}= \frac{{\rm exp}({\bf q}^\top {\bf e}_i)}{\sum\nolimits^k_{j=1} {\rm exp}({\bf q}^\top {\bf e}_j)}, \quad \beta_{i}= \frac{{\rm exp}({\bf q}^\top {\bf t}_i)}{\sum\nolimits^k_{j=1} {\rm exp}({\bf q}^\top {\bf t}_j)}
\end{equation}
where $e_i$ is an entity grounded in the question and $t_i$ is a retrieved timestamp. The candidate entity score and time score are calculated with an aggregated entity ${\bf t}_{att}$ and an aggregated time representation ${\bf t}_{att}$ by the following formulas, respectively.
\begin{equation}
\varphi_{ent}={\bf Re}(\left \langle{\bf e}_{att}, {\bf P}_E{\bf q} \odot {\bf t}_{att}, \overline{\bf e}_{\varepsilon}\right \rangle)
\label{ent_score}
\end{equation}
\begin{equation}
\varphi_{time}={\bf Re}(\left \langle{\bf e}_{att}, {\bf P}_T{\bf q} \odot {\bf t}_{\varepsilon}, \overline{\bf e}_{att}\right \rangle)
\label{time_score}
\end{equation}
$ \overline{\bf e}_{\varepsilon}$ and $ \overline{\bf e}_{\varepsilon}$ are candidate entity and timestamp. The framework of JMFRN is shown in Figure \ref{framework}.

\subsection{Model Training}
To train our JMFRN model, we minimize the binary cross entropy (BCE) loss $L_1$ so that the correct answers get higher scores. The loss of the i-th candidate is calculated as follows.
\begin{equation}
L_1=\sum_q \sum_{c\in C_q} -y_c{\rm log}\sigma(\varphi_c)-(1-y_c){\rm log}(1-\sigma(\varphi_c))
\end{equation}
$C_q$ is the candidate set of question $q$, $y_c$ is the label of candidate $c$. To ensure that question types and answer types are consistent, we design an Answer Types Discrimination (ATD) task as a supplementary. We calculate the probability that the answer type is time and the type loss $L_2$ as follows.
\begin{equation}
p_{time}=\sigma({\bf W}_{type}{\bf q}), \quad \quad L_2= \sum\nolimits_q (z_q-p_{time})^2
\end{equation}
where ${\bf W}_{type}$ is a learnable projection matrix, $\sigma(\cdot)$ sigmoid, $z_q$ is the answer type of question $q$. If the answer is a timestamp $z_q=1$, else $z_q=0$. The joint objective is:
\begin{equation}
L=L_1+\lambda L_2
\label{equation: 14}
\end{equation}
where $\lambda$ is a hyperparameter to adjust the importance of the type discrimination task.

\section{Experiments}
\subsection{Experimental Setting}
\textbf{Dataset.} To validate the effectiveness of our proposed model for the temporal KGQA task, we employ a challenging dataset TimeQuestions which is wildy used \cite{jia2018tempquestions}.\\
\textbf{Baselines.} We selected several non-temporal and temporal models for comparison. Non-temporal models include BERT \cite{kenton2019bert} and traditional KGQA models Uniqorn \cite{pramanik2021uniqorn}, GRAFT-NET \cite{sun2018open} and PullNet \cite{sun2019pullnet}. Temporal models include embedding-based models CRONKGQA \cite{saxena2021question}, TempoQR \cite{mavromatis2022tempoqr}, TMA \cite{TMA} and graph neural network based model EXAQT \cite{jia2018tempquestions}. 

\begin{table}[!t]
\centering
\renewcommand\arraystretch{1.0}
\caption{Hits@1 for different models on TimeQuestions dataset.}
\begin{tabular}{lccccc}
\hline
Models & Ovr & Exp & Imp & Temp & Ord \\ \hline
BERT      & 0.018   & 0.006    & 0.007    & 0.028    & 0.030    \\
PullNet   & 0.105   & 0.022    & 0.081    & 0.234    & 0.029   \\
Uniqorn  & 0.331   & 0.318    & 0.316    & 0.392    & 0.202   \\
GRAFT-NET & 0.452   & 0.445    & 0.428    & 0.515    & 0.322   \\ \hline
CRONKGQA   & 0.393   & 0.388    & 0.380     & 0.347    & 0.457   \\
TMA & 0.436   & 0.442    & 0.419     & 0.476    & 0.352   \\
TempoQR   & 0.459   & 0.503    & 0.442    & 0.485    & 0.367   \\
EXAQT    & 0.572   & 0.568    & 0.512    & 0.642    & 0.420    \\ \hline
JMFRN     & \textbf{0.628}   & \textbf{0.662}    & \textbf{0.530}     & \textbf{0.646}    & \textbf{0.553}   \\ \hline
\end{tabular}
\label{main result}
\end{table}


\subsection{Results and Analysis}
Results of JMFRN and baselines are showed in Table \ref{main result}. \textit{Ovr}, \textit{Exp}, \textit{Imp}, \textit{Temp} and \textit{Ord} means \textit{Overall}, \textit{Explicit}, \textit{Implicit}, \textit{Temporal} and \textit{Ordinal} questions in TimeQuestions separately.

In Table \ref{main result}, firstly, it can be seen that using only the language model BERT hardly works on the TKGQA task. The introduction of KGs will be a great gain in question answering. Secondly, different from traditional KGQA, it is more important for TKGQA reasoning with temporal information. Traditional KGQA models rely on multi-hop reasoning on relation paths in the KG to find the correct answer for a complex question. However, most temporal facts are not connected by relation paths in the TKG, but they are temporally related. Treating TKGQA as a multi-step reasoning problem leads to poor performance, e.g. GRAFT-NET and PullNet in Table \ref{main result}. 

\begin{figure}[ht]
	\centering
	\includegraphics[width=0.8\columnwidth]{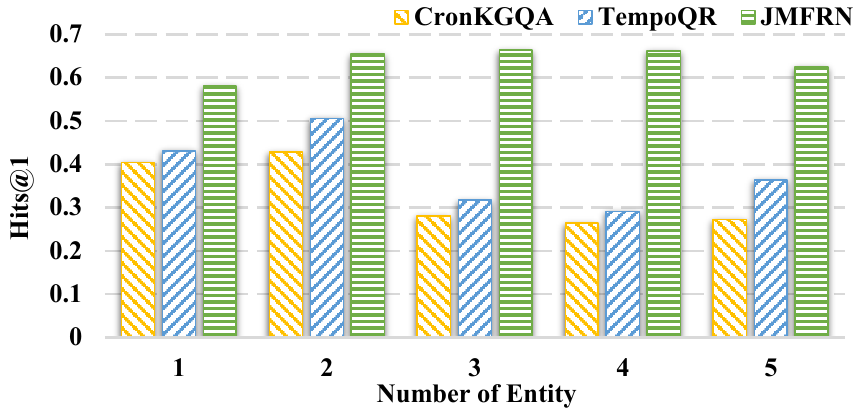}
	\caption{Performance of different numbers of grounded entities.}
	\label{entity_num}
\end{figure}

By contrast, it can be observed that the performance of the temporal models is overall better than that of the non-temporal models. Our proposed JMFRN outperforms the baseline models on all types of questions, especially \textit{Implicit} and \textit{Ordinal}. In Figure \ref{entity_num}, we compare the performance of the TKGE-based model and JMFRN for answering questions containing different numbers of grounded entities. For questions with more than three grounded entities, JMFRN significantly outperforms CRONKGQA and TempoQR. It validates the advantage of JMFRN over previous TKGE-based models in handling complex temporal questions containing multiple entities.

\begin{table}[!ht]
    \centering
    \caption{Ablation results. `w/o' denotes that the module is removed.}
    \begin{tabular}{lccccc}
    \hline
        ~ & Ovr & Exp & Imp & Temp & Ord \\ \hline
        JMFRN & 0.628 & 0.662 & 0.530 & 0.646 & 0.553 \\ \hline
        w/o EAM & 0.591 & 0.633 & 0.519 & 0.575 & 0.547  \\ 
        w/o TAM & 0.617 & 0.641 & 0.511 & 0.645 & 0.550  \\ 
        w/o time fuse & 0.612 & 0.645 & 0.511 &0.629 & 0.545  \\ 
        only EAM & 0.607 & 0.635 & 0.526 & 0.620 & 0.540 \\ \hline
    \end{tabular}
\label{Ablation Results}
\end{table}

\begin{figure}[ht]
	\centering
	\includegraphics[width=0.8\columnwidth]{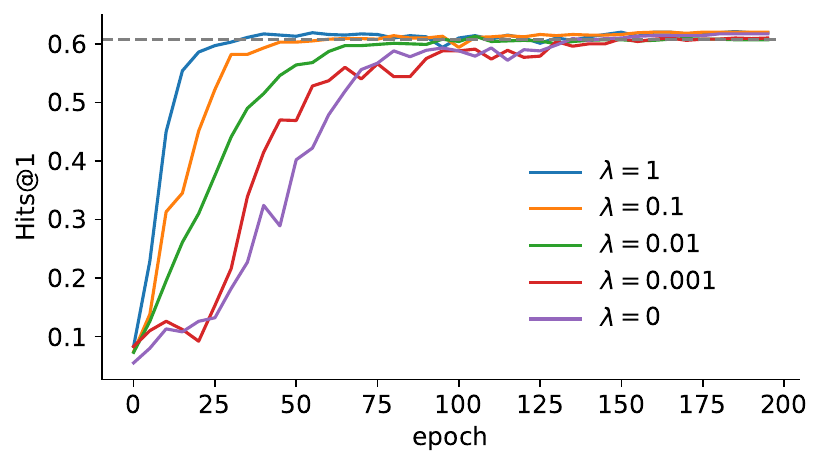}
	\caption{Convergence process of JMFRN with different $\lambda$.}
	\label{convergence}
\end{figure}

\textbf{Ablation Study.} We perform ablation experiments on each components for JMFRN, \ie EAM, TAM and ATD. As shown in Table \ref{Ablation Results}, each component contributes to JMFRN. In particular, the performance of the model drops dramatically to 0.591 after removing the EAM. It validates our motivation that handling multiple entities in complex questions and their question-related facts can improve the accuracy of the Q\&A. we perform further ablation experiments on time information fusion. We first remove temporal fusion during question encoding, and subsequently we remove TAM for inference. Directly removing the time fusion, the performance of the model has declined. Further, removing TAM based on removing time information fusion, i.e., retaining only EAM, the performance of the model decreases to 0.607.\\
\textbf{ATD accelerates and stabilizes the training process.} To filter out mismatched answer types, JMFRN introduces ATD during the training process. In Figure \ref{convergence}, the different $\lambda$ indicate the weight of the ATD task. It can be seen that as the value of $\lambda$ becomes larger, the model converges faster and the convergence process is more stable. We believe that the type discrimination task helps the model to filter a large number of incorrect answers. That is, if the answer type of the question is time, the model focuses on optimizing the time embedding and thus accelerates and stabilizes model convergence.

\section{Conclusion}
It is necessary to answer complex temporal questions by jointly constraining with multiple temporal facts. To handle complex temporal questions, we propose a Joint Multi-Facts Reasoning Network (\textbf{JMFRN}) for Complex Temporal Question Answering over Knowledge Graph. JMFRN starts by retrieving candidate question-relevant temporal facts as a basis. By aggregating entity and temporal information, JMFRN realizes joint multi-fact comprehensive reasoning without specific entity annotations. In addition, the introduction of ADT improves model performance and speed up the convergence of the model. Experiments show that JMFRN achieves state-of-the-art on the complex temporal Q\&A dataset TimeQuestions.



\small
\bibliographystyle{IEEEbib}
\bibliography{main}

\end{document}